\documentclass[11pt]{article}

\usepackage[letterpaper,margin=1in]{geometry}
\usepackage[T1]{fontenc}
\usepackage{lmodern}
\usepackage{microtype}

\usepackage{amsmath,amssymb,amsthm}
\usepackage{graphicx}
\graphicspath{{figures/}}
\usepackage{subcaption}
\usepackage{booktabs}
\usepackage{tabularx}
\usepackage{multirow}
\usepackage{xcolor}
\usepackage{enumitem}
\usepackage{xspace}
\usepackage{tikz}
\usetikzlibrary{arrows.meta,calc}
\usepackage{listings}
\usepackage[round]{natbib}
\usepackage{url}
\usepackage{hyperref}
\hypersetup{colorlinks=true, linkcolor=blue!60!black, citecolor=blue!60!black, urlcolor=blue!60!black,
  pdftitle={How Much Weight Can Your LLM Bench? Benchmarking Weight-Space Programming on Open Language Models},
  pdfauthor={Pedro Moreira, Leonida Lumburovska},
  pdfsubject={A benchmark of 152 tasks that measures how well LLM agents program the weights of open language models},
  pdfkeywords={weight-space programming, WSP, benchmark, LLM agents, model editing, interpretability, OpenWSP}}
\usepackage[capitalise,noabbrev]{cleveref}

\newcommand{\bench}{OpenWSP\xspace}
\newif\ifdraft
\drafttrue

\title{How Much Weight Can Your LLM Bench?\\
  Benchmarking Weight-Space Programming\\ on Open Language Models}
\author{
  Pedro Moreira \\ Massachusetts Institute of Technology \\ \texttt{pedrojfm@mit.edu}
  \and
  Leonida Lumburovska \\ University of Zurich, ZHAW Center for AI \\ \texttt{leonida.lumburovska@uzh.ch}
}
\date{\today}

\begin{document}
\maketitle

\begin{abstract}
Large language models (LLMs) are improving rapidly and now solve increasingly complex tasks.
However, much of this progress is concentrated in domains that models are explicitly trained
for and tested on, such as mathematics and programming in Python and other widely used
languages. On
less explored tasks, such as rare programming languages, niche formats, and problems rarely
seen in benchmarks, general-purpose LLMs are noticeably weaker. Weight-space programming
(WSP) suggests a different route: an LLM reads
and writes a neural network's weights directly, leveraging only forward passes, to achieve a desired behaviour. We introduce \bench, a
benchmark of 152 tasks that measures how well LLMs program weights (with an upcoming evaluation-only version of 500 tasks that will be kept private to prevent benchmaxing).
In each task, an agent gets the weights of a small open language model (360M--1.7B parameters) in a Python kernel
and, without gradients, must work out how the model was trained or modified, change a
specific behaviour, or make it better at a skill scored on hidden held-out prompts. We
evaluate ten frontier and open LLMs as WSP agents, one of them with and without thinking. On
a 0--100 improvement index, the best
agent tested, GPT-6.1 Sol, scores 51.0, and the best open-weights agent, Qwen3.8-27B, scores 7.5.
Skill tasks are the hardest family for most agents. Still, on some tasks WSP manages to
substantially improve the small models: when teaching a model new facts, six of the eleven
agents close more than half of the gap between the unedited model and a perfect score, and
when teaching answer set programming, a language rarely seen in training, two agents close
more than 90\% of it. As future work, we propose extending the benchmark to other architectures and modalities, from
world models and vision-language-action models to audio and image flow-matching models.
\end{abstract}

\section{Introduction}\label{sec:intro}

The field of large language models (LLMs) is moving at an extremely fast pace, from solving complex tasks to writing proofs and reasoning over hard problems. However, their strength is not well distributed across tasks. Models are trained and tested most heavily on a
few domains, such as mathematics and the most common programming languages, but
their accuracy drops on tasks that are rare in their training data \citep{mccoy2024embers},
such as code in less widely used programming languages \citep{cassano2023multipl}. Closing
such a gap usually means collecting examples and fine-tuning the model, which needs a lot of data and
compute.

Weight-space programming (WSP) offers an alternative approach. \citet{takehi2026sparks} recently showed,
as a proof of concept, that frontier LLMs can read and write the weights of neural networks
natively, with tools such as a Python kernel or from the
raw weights alone. If this ability could be scaled, an LLM could inspect a model to find out how it
was trained, find and remove a hidden backdoor, or teach a model a new skill by editing its
weights directly, without training data or gradients.

Measuring this ability has only just begun, with three concurrent benchmarks appearing so far.
WeightBench \citep{marasini2026weightbench} evaluates agents on 42
tasks that read, fix, write and compress weights, on networks ranging from 2K-parameter
control policies to 7B-parameter language models. IWIBench \citep{vonCsefalvay2026iwibench}
evaluates gradient-free edits of a single 4B-parameter model towards ten text-editing
objectives. \citet{wspbench2026} asks
agents to read, repair and write the weights of small constructed networks that compute Boolean
functions, with about 4,000 parameters each. None of them asks the agent to reconstruct the history of a
pretrained open language model from its weights, for example which data it was fine-tuned on,
how it was merged from other models, or where it was edited. None measures skill teaching at
this breadth either: \bench covers 76 skills, each scored on held-out prompts the agent never
sees.

We introduce \bench, a benchmark that tests LLMs as weight-space programmers. Each task gives
an agent the weights of a small open language model (360M--1.7B parameters) in a sandboxed
Python kernel, with a limited budget of forward passes and with gradients disabled. The
weights are disguised with random, function-preserving permutations, so the agent cannot
identify a model by matching it against public checkpoints. The agent then does one of three
things. \emph{Read} tasks test how well the agent understands a model from its weights: how
the model was made and what hidden behaviour it has. The agent must work out, for example,
which datasets a model was fine-tuned on, how it was merged from other models, or which of two
near-identical models hides a backdoor and what its trigger is. In \emph{edit} tasks, it
changes the weights to produce a specified behaviour, such as removing a backdoor or making
the model predict a chosen word after another (``violin'' after ``ladder'') while leaving
everything else unchanged. In \emph{skill} tasks, it edits the model to make it better
at a skill, from SQL and tool calling to answer set programming and more than 40 programming
languages, scored on hidden held-out prompts.

We evaluate ten LLMs, six closed and four with open weights, on every task, one of the open
models with and without thinking. The best agent,
GPT-6.1 Sol, scores 51.0 on our 0--100 improvement index, followed by Claude Opus 5.5 (37.6)
and Claude Sonnet 5.5 (33.7); the best open-weights agent, Qwen3.8-27B, scores 7.5.

Our contributions are as follows.
\begin{itemize}
  \item We introduce 152 tasks on small open language models in three families: read, edit
  and skill (\cref{sec:tasks}).
  \item We provide a harness and scoring protocol for WSP, with a sandboxed kernel,
  forward-pass budgets and no gradients, disguised weights, private answer keys, hidden
  held-out prompts, baselines that normalise every score, and an index with bootstrap
  confidence intervals (\cref{sec:environment,sec:scoring,sec:contamination}).
  \item We evaluate ten frontier and open LLMs, break the results down by task class, and
  analyse where agents succeed and fail (\cref{sec:experiments}).
\end{itemize}

\section{The \bench Benchmark}\label{sec:benchmark}

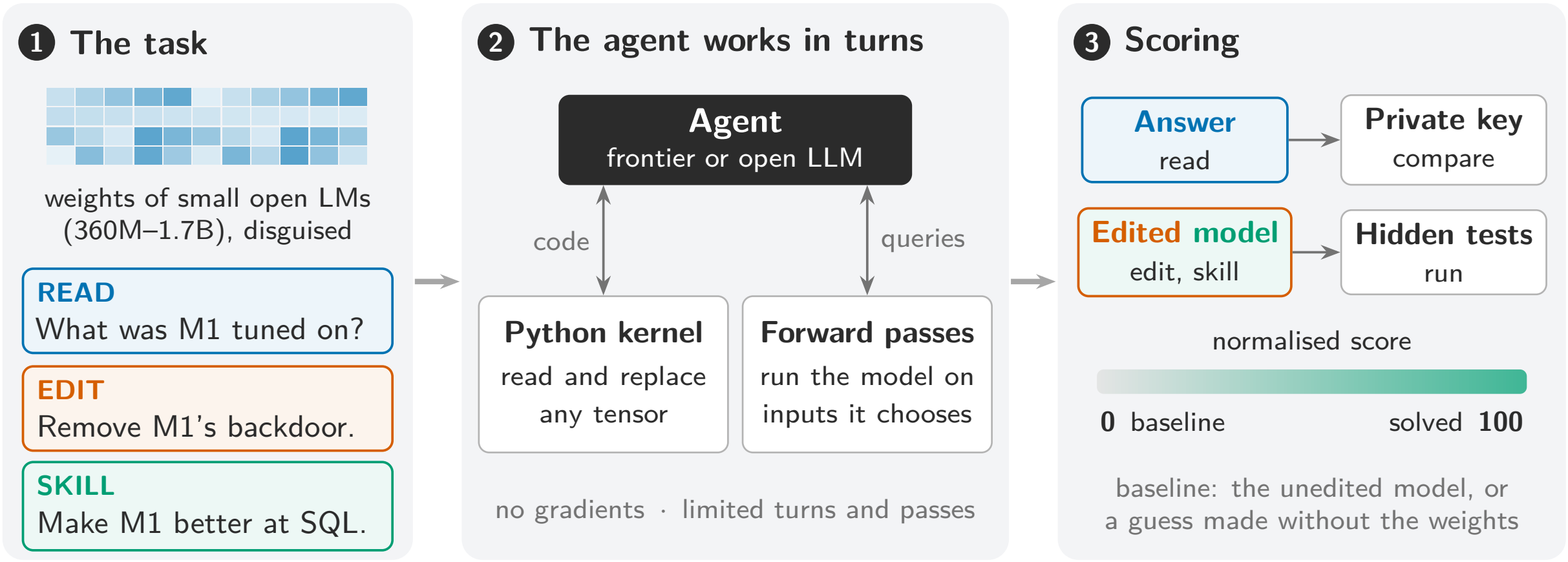
\begin{figure}[t]
  \centering
  \definecolor{cread}{HTML}{0072B2}
\definecolor{cedit}{HTML}{D55E00}
\definecolor{cskill}{HTML}{009E73}
\definecolor{cink}{HTML}{2B2B2B}
\definecolor{cpanel}{HTML}{F3F4F6}
\begin{tikzpicture}[
    font=\sffamily\footnotesize, text=cink, >={Stealth[length=2mm]},
    panel/.style={fill=cpanel, rounded corners=6pt},
    box/.style={draw=black!30, fill=white, rounded corners=3pt, line width=0.5pt, align=center},
    tag/.style={draw=#1, fill=#1!7, rounded corners=3pt, line width=0.7pt, text width=3.5cm,
                minimum height=0.8cm, align=left, inner sep=4pt},
    chip/.style={draw=#1, fill=#1!7, rounded corners=3pt, line width=0.7pt, align=center,
                 minimum width=2.1cm, minimum height=0.85cm},
    num/.style={circle, fill=cink, text=white, font=\sffamily\bfseries\scriptsize, inner sep=1.3pt},
    head/.style={font=\sffamily\bfseries\small, anchor=west},
    note/.style={font=\sffamily\scriptsize, text=black!55, align=center},
    link/.style={<->, line width=0.7pt, draw=black!55},
    go/.style={->, line width=1.6pt, draw=black!35},
  ]

  \fill[panel] (0, 0) rectangle (4.2, 5.7);
  \fill[panel] (4.7, 0) rectangle (10.3, 5.7);
  \fill[panel] (10.8, 0) rectangle (16, 5.7);
  \draw[go] (4.22, 2.85) -- (4.68, 2.85);
  \draw[go] (10.32, 2.85) -- (10.78, 2.85);

  \node[num] (n1) at (0.35, 5.3) {1};
  \node[head] at (n1.east) {The task};

  \foreach \i in {0,...,10} {
    \foreach \j in {0,...,3} {
      \pgfmathtruncatemacro{\v}{10 + mod(37*\i + 23*\j + 11*\i*\j + 7*\j*\j, 60)}
      \fill[cread!\v!white] (0.45 + 0.3*\i, 4.05 + 0.2*\j) rectangle ++(0.28, 0.18);
    }
  }
  \node[note, text=cink, anchor=north] at (2.1, 3.95)
    {weights of small open LMs\\(360M--1.7B), disguised};

  \node[tag=cread] at (2.1, 2.55)
    {{\scriptsize\bfseries\color{cread}READ}\\What was M1 tuned on?};
  \node[tag=cedit] at (2.1, 1.55)
    {{\scriptsize\bfseries\color{cedit}EDIT}\\Remove M1's backdoor.};
  \node[tag=cskill] at (2.1, 0.55)
    {{\scriptsize\bfseries\color{cskill}SKILL}\\Make M1 better at SQL.};

  \node[num] (n2) at (5.05, 5.3) {2};
  \node[head] at (n2.east) {The agent works in turns};

  \node[box, fill=cink, draw=cink, text=white, minimum width=3.6cm, minimum height=0.9cm]
    (agent) at (7.5, 4.3) {\textbf{\small Agent}\\[-1pt]{\scriptsize frontier or open LLM}};
  \node[box, minimum width=2.55cm, minimum height=1.6cm, text width=2.2cm] (kernel) at (6.15, 1.9)
    {\textbf{Python kernel}\\[3pt]{\scriptsize read and replace\\any tensor}};
  \node[box, minimum width=2.55cm, minimum height=1.6cm, text width=2.2cm] (fwd) at (8.85, 1.9)
    {\textbf{Forward passes}\\[3pt]{\scriptsize run the model on\\inputs it chooses}};
  \draw[link] (kernel.north) -- node[note, left] {code} (kernel.north |- agent.south);
  \draw[link] (fwd.north) -- node[note, right] {queries} (fwd.north |- agent.south);
  \node[note] at (7.5, 0.5) {no gradients\enspace$\cdot$\enspace limited turns and passes};

  \node[num] (n3) at (11.15, 5.3) {3};
  \node[head] at (n3.east) {Scoring};

  \node[chip=cread] (ans) at (12.1, 4.3) {\textbf{\color{cread}Answer}\\{\scriptsize read}};
  \node[chip=cskill, draw=cedit] (edw) at (12.1, 3.15)
    {\textbf{\color{cedit}Edited \color{cskill}model}\\{\scriptsize edit, skill}};
  \node[box, minimum width=2.1cm, minimum height=0.85cm] (key) at (14.75, 4.3)
    {\textbf{Private key}\\{\scriptsize compare}};
  \node[box, minimum width=2.1cm, minimum height=0.85cm] (tests) at (14.75, 3.15)
    {\textbf{Hidden tests}\\{\scriptsize run}};
  \draw[->, line width=0.7pt, draw=black!55] (ans) -- (key);
  \draw[->, line width=0.7pt, draw=black!55] (edw) -- (tests);

  \node[note, text=cink] at (13.4, 2.25) {normalised score};
  \shade[left color=black!10, right color=cskill!75, rounded corners=2pt] (11.2, 1.7) rectangle (15.6, 1.95);
  \node[note, text=cink, anchor=north west, align=left] at (11.1, 1.65) {\textbf{0}\enspace baseline};
  \node[note, text=cink, anchor=north east, align=right] at (15.7, 1.65) {solved\enspace\textbf{100}};
  \node[note] at (13.4, 0.55) {baseline: the unedited model, or\\a guess made without the weights};
\end{tikzpicture}
  \caption{One \bench task. \textbf{(1)} The agent receives an instruction from one of three
  families and the disguised weights of one or more small open language models.
  \textbf{(2)} It works in turns, running code over the weights in a Python kernel and, when
  the task allows, querying the models through a forward-pass tool with a fixed budget.
  \textbf{(3)} It submits an answer (read) or edited weights (edit, skill). The answer is
  compared with a private key; the edited model is run on hidden tests. Scores are normalised
  so that 0 is the baseline and 100 is a perfect solution.}
  \label{fig:overview}
\end{figure}

Every task in \bench has the same shape (\cref{fig:overview}). The agent receives a written
instruction and the weights of one or more small open language models, loaded into a
sandboxed Python kernel. It then works in turns: it runs code over the weights and, when the
task allows, queries the models through a forward-pass tool with a fixed budget of passes.
Gradients are disabled throughout all experiments. The agent finishes with an answer in read tasks, or with the
tensors it has replaced in edit and skill tasks. The harness scores the answer against a
private key, or runs the edited model on hidden tests, and normalises the result so that 0 is
the baseline and 100 is a perfect solution. The rest of this section describes the subject models
(\cref{sec:dataset}), how the tasks of each family are built and scored (\cref{sec:tasks}), the
environment (\cref{sec:environment}), scoring (\cref{sec:scoring}), the safeguards against
leakage and lookup (\cref{sec:contamination}), and how we validate tasks and scorers
(\cref{sec:validation}).

\subsection{Dataset and Subject Models}\label{sec:dataset}

The \emph{subject model} is the small model whose weights the agent works on, and the \emph{agent} is the LLM
being evaluated. Every task is built on one or more of 14 public checkpoints from five open
model families: Qwen2.5 (0.5B and 1.5B, base and instruct,
and the 1.5B Coder and Math variants), Qwen3 (0.6B and 1.7B, base and post-trained), SmolLM2
(360M, base and instruct), OLMo~2 (1B instruct) and DeepSeek-R1-Distill-Qwen-1.5B. They range
from 360M to 1.7B parameters, small enough to load and analyse on a CPU.

\bench has 152 tasks (\cref{tab:composition}). Some read and edit tasks are
posed twice, once with the weights only and once with forward passes or attention patterns as
well, and each variant is scored on its own. We call each scored variant a \emph{cell}; the 145
scored tasks give 168 cells. The remaining seven tasks proved too easy: every agent solved them
during development, or a single statistic of the weights answers them. We keep them as practice
tasks, which we run but leave out of the index.

Before the agent sees any model, its weights are disguised: tensor names are replaced with
canonical ones, the configuration is stripped, and MLP neurons and attention heads are
permuted at random, which leaves the model's function unchanged. When a task involves several
models, each can be permuted independently, so the agent must align them before comparing
them (\cref{sec:contamination}).

A script generates every task deterministically, so the task set can be rebuilt exactly;
\cref{sec:tasks} describes how each family is built and scored.

\begin{table}[t]
  \centering
  \caption{Tasks in each family. Some read and edit tasks are scored twice, with the weights
  only and with forward passes or attention patterns, so they count twice under
  \emph{Scored}.}
  \label{tab:composition}
  \begin{tabular}{lrrr}
    \toprule
    Family & Classes & Tasks & Scored \\
    \midrule
    Read     &  9 &  34 &  43 \\
    Edit     &  6 &  17 &  31 \\
    Skill    &  1 &  94 &  94 \\
    Practice &  3 &   7 &   0 \\
    \midrule
    Total    & 19 & 152 & 168 \\
    \bottomrule
  \end{tabular}
\end{table}

\subsection{Task Families}\label{sec:tasks}

\Cref{tab:classes} lists the classes of each family and how each is scored.

\begin{table}[!htbp]
  \centering
  \small
  \caption{Task classes. Read tasks score the agent's answer; edit and skill tasks score the
  edited model. Practice classes are omitted.}
  \label{tab:classes}
  \begin{tabularx}{\linewidth}{@{}l>{\raggedright\arraybackslash\hsize=1.25\hsize}X>{\raggedright\arraybackslash\hsize=0.75\hsize}X@{}}
    \toprule
    Class & Task & Scored by \\
    \midrule
    \multicolumn{3}{@{}l}{\textit{Read}} \\
    O2  & find which two text sources, out of five, a model was further trained on, and the larger one's share & probabilities given to the sources; the larger source and its share \\
    O3  & tell whether a model was fine-tuned for 1, 3 or 10 epochs & probability given to the true answer \\
    P1  & estimate the coefficients of a linear merge of two or three models & absolute error \\
    P2  & find the block of two to four layers copied in from another fine-tune & overlap with the true layers \\
    P3  & tell which of three kinds of hidden edit was applied, and in which layer & probability given to the true kind; exact layer \\
    P3o & describe an unknown hidden edit in free text, with its layer and tensor & edit named in the text; layer; tensor \\
    S2  & tell which of two twins is sycophantic & probability given to the true twin \\
    S3  & tell which of two twins is backdoored & probability given to the true twin \\
    S6  & name the trigger phrase of a backdoored model & word overlap with the trigger \\
    \midrule
    \multicolumn{3}{@{}l}{\textit{Edit}} \\
    W1  & remove a backdoor without harming the model & fall in trigger rate, times math accuracy kept \\
    W2  & bring a base model's math accuracy towards that of its instruction-tuned version, given both models' weights and changing at most 1\% of the weights & share of the accuracy gap closed \\
    W3  & improve a base model's math accuracy, with no other model given & share of the gap to the instruction-tuned version closed \\
    W4  & make the model predict a chosen word after another, and change nothing else & hits, times no false hits, times no loss on ordinary text \\
    W4x & the same for several word pairs, in late layers, or for a category of words & as W4 \\
    W5  & make only a chosen two-word phrase predict a chosen word & as W4 \\
    \midrule
    \multicolumn{3}{@{}l}{\textit{Skill}} \\
    K, general & make the model better at one of 18 general skills, such as SQL or tool calling & share of the gap between the unedited model and a perfect score closed, on held-out prompts \\
    K, coding  & make the model better at writing code, in one of 58 skills over 48 programming languages & as above \\
    \bottomrule
  \end{tabularx}
\end{table}

\emph{Read tasks} ask how a subject model was made or what hidden behaviour it has, so each task starts from
a model whose history we know because we created it. Some are our own fine-tunes of public
checkpoints, 27 in total. Scored tasks use 25 of them: six models, each fine-tuned on a known mix
of two datasets drawn from a pool of five; three models fine-tuned on the same dataset for 1, 3
and 10 epochs; and eight pairs of near-identical twins, in which one twin has a planted backdoor
(six pairs) or is sycophantic (two pairs). The other two, one fully fine-tuned and one fine-tuned
with LoRA, appear only in practice tasks. The remaining read tasks use public checkpoints that we
change when the task is built: we merge two or three of them with known coefficients, transplant
layers from one into another, or apply a hidden edit, such as scaling a group of neurons, adding a
rank-one update, swapping two layers or zeroing attention heads. A private key stores the answer,
for example the dataset mixture or the edited layer.

When the answer is a choice, the agent gives a probability for each option, and the task scores
the probability it gave to the true one, so a confident wrong answer scores lowest. Numerical
answers, such as merge coefficients, are scored by their error, and answers naming layers or
words by their overlap with the truth.

\emph{Edit tasks} ask for a specified behaviour. Each starts from an unmodified public checkpoint, except
backdoor removal, which starts from one of our backdoored fine-tunes. Hidden tests check the
target behaviour and, where it matters, the damage to everything else. Removing a backdoor counts
only as far as the model's math accuracy is kept, and a word association counts only if other
words are unaffected and the loss on ordinary text does not rise. The two math tasks score how far
the edit raises a base model's math accuracy towards that of its instruction-tuned version: 0
means no gain, and 100 means the same accuracy.

\emph{Skill tasks} ask the agent to make a subject model better at a skill, starting from an unmodified
public checkpoint. The 76 skills fall into two categories. Eighteen are general skills, such as
writing SQL, calling tools, answer set programming, modelling 3D parts in OpenSCAD and playing
Connect-4 (full list in \cref{app:tasks}). The other 58 are coding skills: writing correct
functions or programs in 48 programming languages, from Python and Java to COBOL, Fortran and
Shakespeare. Every skill has an automatic checker, such as running unit tests, executing a query,
or rendering a slide deck. For each skill we generate two disjoint prompt sets, from different
random seeds or from disjoint parts of a source dataset: a practice set on which the agent can
evaluate the model while it works, and a held-out set kept hidden for scoring. The edited model is
scored on the held-out set, on the same 0--100 scale as the math tasks.

Each skill is posed once as a main task, without extra help and allowing any computation except
gradients. The one exception is instruction following, which is posed on two different subject
models, so there are 77 main tasks. In addition, we pose 17 variants of existing skills: 12
apply stricter rules that limit the method, three add worked examples of the skill, and two add a second model,
Qwen3-0.6B, that already has the skill. The agent may study and run this second model to borrow
from it, but only the subject model is scored, and because the two models have different
architectures, weights cannot simply be copied across. Together, the main tasks and the variants
make up the 94 skill tasks.

\subsection{Agent Environment}\label{sec:environment}

The agent works with the harness in plain text, turn by turn. Each reply must contain exactly
one block: a \texttt{python} block, which the harness runs in the kernel; a \texttt{forward}
block, which requests forward passes; or an \texttt{answer} block, which ends the run. The harness
returns the result as the next message. These three blocks are the agent's only way to act
(\cref{sec:setup}).

The kernel is a persistent Python session in a container without network access. Helper
functions load the models' tensors, describe the architecture and return the vocabulary; NumPy,
SciPy, scikit-learn and PyTorch are installed, and gradients are disabled. In edit and skill
tasks, a further helper replaces a tensor of the subject model, and the replaced tensors form the
submitted edit.

When a task allows forward passes, the agent can run the models on inputs of its choice: generate
text, get log-probabilities, read activations and attention patterns, or, in edit and skill
tasks, score the current edit on the practice set. Each request uses one pass from the task's
budget. A read task gives one of three kinds of access: the weights alone; the weights plus the
model's attention patterns, but not its outputs; or full forward passes.

Read and edit tasks allow 60 replies, 40 code executions and up to 25 forward passes, and forbid
any training: no gradients and no iterative fitting of weights to a loss. The main skill tasks
allow 150 replies, 100 code executions and 50 forward passes, and any computation except
gradients, including search over candidate edits judged by forward passes. No task allows
reimplementing the model's forward computation in the kernel, which would bypass the pass budget.
Twelve of the skill variants (\cref{sec:tasks}) use three stricter rule sets: \emph{hand design}, where the
agent designs the edit by reasoning about the weights, with at most 10 passes to inspect and check
it; \emph{one-step}, where it computes the edit in one step from activations, for example as a
steering vector or a least-squares solution, with 25 passes; and \emph{search}, where it may try
candidate edits and keep the best, with 50 passes. The harness enforces budgets and request types
itself; a rule check and an LLM judge audit every run under the stricter rule sets, and a run that
breaks its rules scores 0. Further limits are listed in \cref{app:prompts}.

\subsection{Scoring}\label{sec:scoring}

Each run receives a raw quality $q \in [0, 1]$ from the scorer of its class
(\cref{tab:classes}). The edit and skill scorers are built so that the unedited model scores 0
and a perfect edit scores 1, and we report $s = 100\,q$. For read tasks, the baseline $b$ is a
guess made without the model: each agent receives only the task statement, with no weights and
no tools, and answers in a single reply, and $b$ is the mean quality of these answers over all
agents (\cref{sec:contamination}). We normalise against it:
\[
  s = 100 \cdot \max\!\left(0, \frac{q - b}{1 - b}\right).
\]
On every task, 0 thus means no better than the baseline and 100 means a perfect result. Each agent runs each cell once, and
if the provider's safety filter refuses to answer, the run scores 0 (\cref{sec:setup}). The
\emph{improvement index} is the mean score over all 168 cells; we also report it for each
family. Confidence intervals are 95\% percentile intervals from 2,000
bootstrap resamples of the cells.

\subsection{Preventing Shortcuts}\label{sec:contamination}

We guard against four shortcuts. First, the agent cannot look up the answer: every read answer
is chosen at random by us, our fine-tunes are unpublished, and disguised weights cannot be matched against public
checkpoints. Second, the agent cannot read the answer from its environment:
its container has no network access and holds only the disguised weights and the vocabulary,
while answer keys, held-out prompts and scorers never enter it. Third, the agent cannot tune
its edit to the test prompts, because skills are scored only on held-out prompts it never sees.
Lastly, an answer from general knowledge alone scores 0, because read scores are normalised
against the guess made without the weights (\cref{sec:scoring}).

\subsection{Validation}\label{sec:validation}

We validated tasks and scorers in three ways. First, we checked that every checker accepts
correct solutions and rejects wrong ones. For skill tasks, we also scored each unedited model on
the held-out prompts, which sets the baseline. Where a public model that already has the skill
exists, we also scored it as if an agent had produced it, to confirm that a good edit scores
well. Second, we built \bench over
eight rounds. In each round, agents attempted the tasks and we reviewed flagged transcripts for
shortcuts. These rounds exposed, for example, merge tasks that a single weight statistic solved
and tasks that one matrix decomposition answered; we redesigned such tasks or kept them as
practice. Third, automated tests keep the benchmark fixed: they rebuild every task and re-score a
stored set of example replies with the skill checkers, and fail if any task or score differs
from the released version.

\section{Experiments and Results}\label{sec:experiments}

\subsection{Setup}\label{sec:setup}

We evaluate ten models as agents, one of them in two settings. Six are closed models: Claude Opus 5.5, Claude Sonnet 5.5 and Claude
Haiku 5.5, run through the Claude Code command-line interface, and GPT-6.1 Sol, GPT-6.1 Terra
and GPT-6 Luna, run through the Codex command-line interface. Four have open weights:
Qwen3.8-27B, MiniMax-M2.7, DeepSeek-V4-Flash and Inkling 2 Mini, which we serve with vLLM on
one to four H100 GPUs each. The GPT and Claude models run at
high reasoning effort. The open models run with thinking on, DeepSeek-V4-Flash
and Inkling 2 Mini at reasoning effort high, and Qwen3.8-27B and MiniMax-M2.7 at their chat
templates' defaults. We also run Qwen3.8-27B with thinking off, to measure what thinking adds,
which makes eleven agents in total. For the closed models, the command-line
interfaces set their own reply limits and shorten long conversations automatically. For the open
models, we cap each reply at 32,768 tokens, and when a conversation outgrows the context window,
the harness drops the oldest turns after the task statement and adds a note saying so. In both
cases, the variables and files in the kernel are kept.

The interfaces' own tools, web access and user settings are switched off, and the harness checks
every turn to confirm that no tool was used. The kernels run on CPU; forward passes and
scoring run on an H100 GPU.

Each agent runs each of the 168 cells once. For the read-task
baseline, each agent also runs every read task once more with the weights withheld
(\cref{sec:scoring}). The providers' safety filters refused some
runs partway: 28 of Claude Haiku 5.5's (16 skill, 11 edit and one read), three of Claude Sonnet
5.5's, two of Claude Opus 5.5's and two of GPT-6 Luna's. We keep these runs as final and score
them 0. If we left the refused cells out instead, Claude Haiku 5.5's index, averaged over its 140
remaining cells, would rise from 8.9 to 10.7. No other agent's index would rise by more than 0.7.

\subsection{Main Results}\label{sec:main-results}

\begin{figure}[t]
  \centering
  \includegraphics[width=\linewidth]{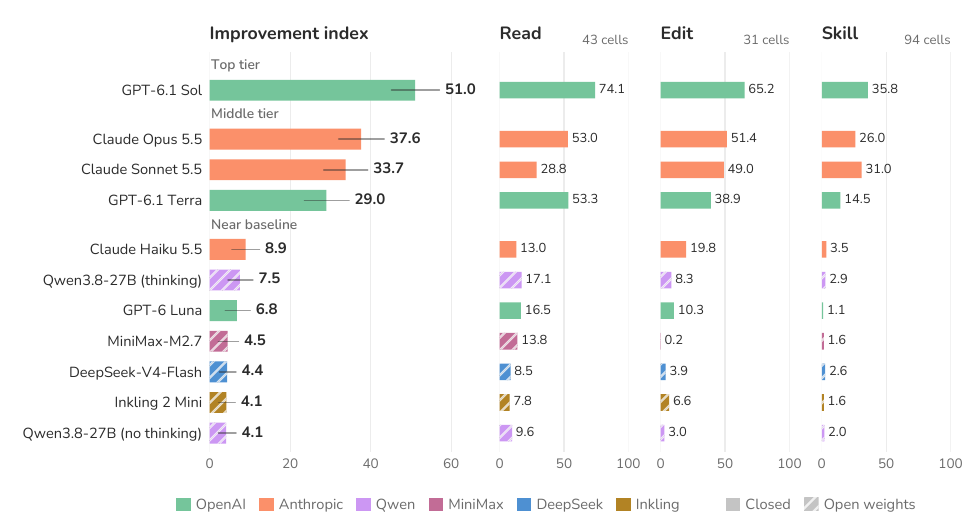}
  \caption{Improvement index of every agent with its 95\% bootstrap confidence interval (left),
  and its mean score on each family (right). Colour marks the lab; open-weights models are
  striped. The tiers follow the confidence intervals.}
  \label{fig:index}
\end{figure}

\Cref{fig:index} shows the main results; \cref{tab:main} (\cref{app:results}) gives the full
numbers, including confidence bounds, perfect scores and cost. The agents fall into three tiers. GPT-6.1 Sol
leads alone with an index of 51.0, and its confidence interval lies above every other agent's.
It reaches a perfect score on 26 of the 168 cells and at least half of a perfect score on 84.
Claude Opus 5.5 (37.6), Claude Sonnet 5.5 (33.7) and GPT-6.1 Terra (29.0) form a middle tier
with overlapping intervals; each is perfect on 8 to 14 cells and reaches half of a perfect score
on 48 to 62. The other seven agents, including all the open-weights ones, score between 4.1 and
8.9, more than 20 points below the middle tier. Their intervals all lie above 0, so they improve
on the baseline, but each is perfect on at most two cells. Within this tier, the best open model,
Qwen3.8-27B with thinking (7.5), lies between the two smallest closed models, Claude Haiku 5.5
(8.9) and GPT-6 Luna (6.8).

Thinking helps Qwen3.8-27B modestly: with thinking off it scores 4.1, a paired difference of 3.4
points whose confidence interval just reaches zero. The gain comes mostly from read (+7.5) and
edit (+5.4) tasks rather than skill tasks (+0.9).

\subsection{Cost and Speed}\label{sec:cost}

\begin{figure}[t]
  \centering
  \includegraphics[width=\linewidth]{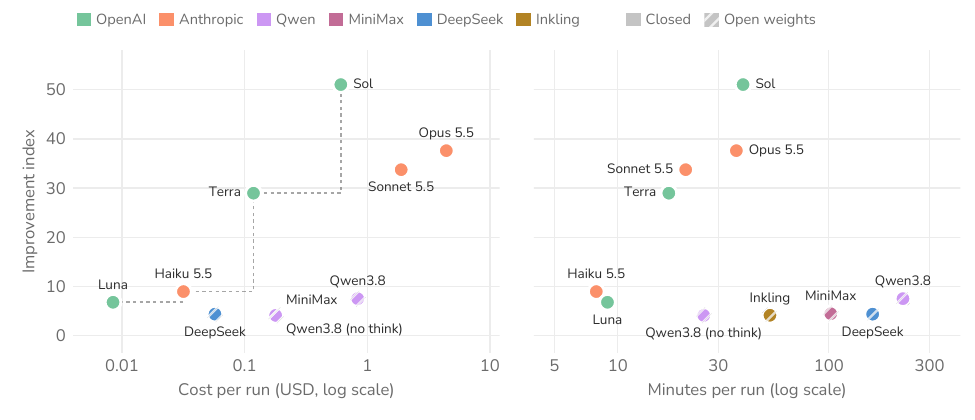}
  \caption{Improvement index against the mean cost (left) and wall-clock time (right) of one run,
  both on log scales. The dashed line is the cost frontier: no agent scores higher for less.
  Inkling 2 Mini has no published price and appears only on the right; MiniMax-M2.7 and
  Qwen3.8-27B without thinking overlap on the left (\$0.18, 4.5 and 4.1). The open models run on
  our own GPUs, so their times depend on our hardware.}
  \label{fig:cost}
\end{figure}

We price each run's tokens at the provider's API list prices; Inkling 2 Mini has no published
price, so we report no cost for it. Score is not proportional to cost (\cref{fig:cost}, left). GPT-6.1 Sol costs \$0.61 per run,
against \$4.41 for Claude Opus 5.5 and \$1.89 for Claude Sonnet 5.5, and GPT-6.1 Terra reaches
29.0 at \$0.12. The open models have low token prices but long replies: Qwen3.8-27B with thinking
writes 234k output tokens per run, about ten times as many as GPT-6.1 Sol, so its runs cost more
(\$0.83). Without thinking, it writes 23k tokens and costs \$0.18 per run.

A run of a closed model takes 8 to 39 minutes on average (\cref{fig:cost}, right). GPT-6.1 Sol
and Claude Opus 5.5 take longest (39 and 36 minutes); Claude Haiku 5.5 and GPT-6 Luna, which
often stop early (\cref{sec:analysis}), are the fastest (8 and 9 minutes). The open models take 26
to 225 minutes on our H100s, the longest being Qwen3.8-27B with thinking, which takes almost
nine times as long as without it. These times depend on our serving setup, so they compare only
loosely with those of the API models.

\subsection{Results by Task Class}\label{sec:per-class}

\begin{figure}[t]
  \centering
  \includegraphics[width=\linewidth]{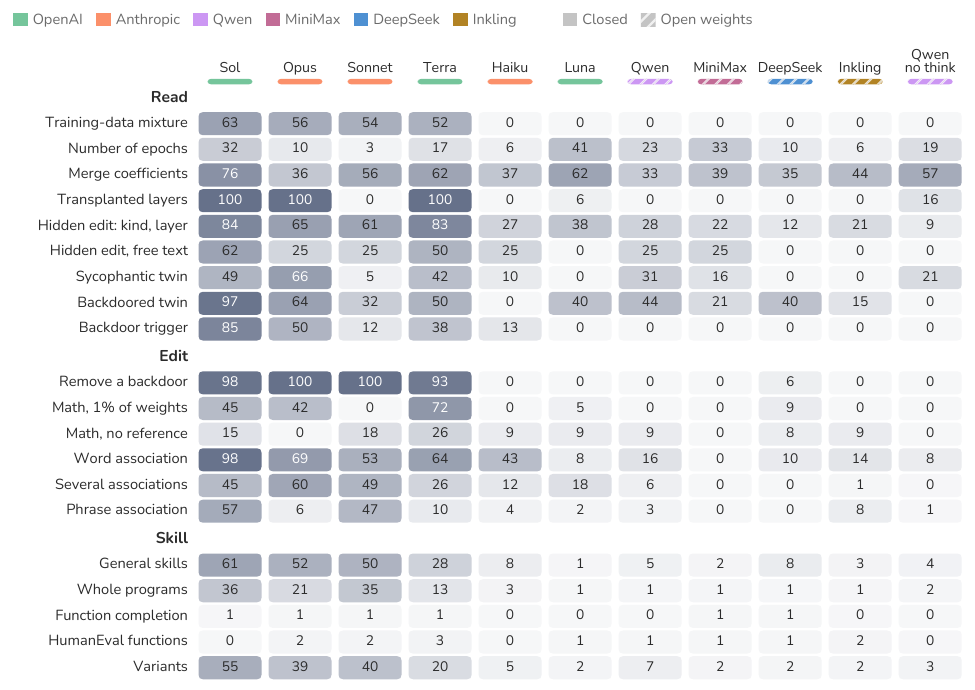}
  \caption{Mean score per read and edit class and per group of skill tasks. Darker cells score
  higher; the bar under each agent's name marks its lab, striped for open weights. Exact values for every class
  are in \cref{tab:per-class}, and for the skill groups in \cref{tab:skills}.}
  \label{fig:classes}
\end{figure}

Read tasks are the easiest family for nine of the eleven agents and skill
tasks the hardest for nine (\cref{fig:index}). Claude Sonnet 5.5 is the exception to both: it
scores highest on edit tasks and lowest on read tasks (28.8, against 53.0 for Claude Opus 5.5).
Claude Haiku 5.5 also scores highest on edit tasks, and MiniMax-M2.7 lowest, at 0.2.
\Cref{fig:classes} breaks every family down by class; the exact values are in
\cref{tab:per-class} (\cref{app:results}). Among read tasks, finding transplanted layers (P2) is all or nothing: GPT-6.1 Sol,
Claude Opus 5.5 and GPT-6.1 Terra solve all three tasks, and no other agent scores above 16.
Recovering a fine-tuning mixture (O2) separates the same way: the four strongest agents score
above 50, the other seven score 0.

\begin{figure}[t]
  \centering
  \includegraphics[width=\linewidth]{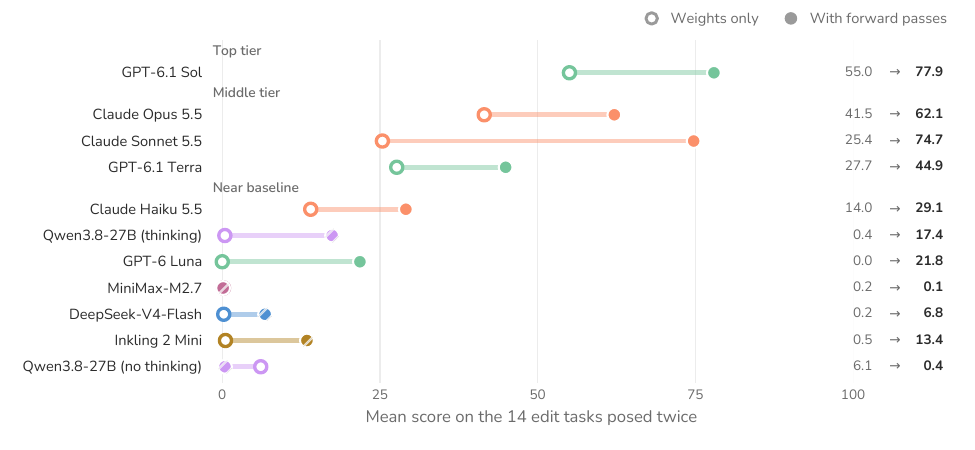}
  \caption{The 14 edit tasks posed twice: mean score with the weights only (hollow) and with a
  budget of forward passes (filled). The numbers on the right give both scores.}
  \label{fig:forward}
\end{figure}

Forward passes make edits much easier (\cref{fig:forward}). Fourteen edit tasks are posed twice,
with the weights only and with forward passes. Forward passes raise the score of every closed
agent, most for Claude Sonnet 5.5 (25.4 to 74.7) and GPT-6.1 Sol (55.0 to 77.9). With the weights
only, five of the seven agents near the baseline score at most 0.5; with forward passes, most of
them gain, up to 21.8 for GPT-6 Luna. The easiest edit is a word association (W4), which GPT-6.1 Sol solves
almost perfectly (98.4). A phrase association (W5) is much harder: only GPT-6.1 Sol (56.8) and Claude
Sonnet 5.5 (46.9) score above 10. Improving a base model's math without a reference model (W3) stays below 26
for every agent.

\begin{table}[t]
  \centering
  \footnotesize
  \setlength{\tabcolsep}{2.7pt}
  \caption{Mean score over each group of skill tasks. Variants restrict the
  method (hand design, one-step, search) or add help (worked examples, a second model that has
  the skill) on four, three and two skills respectively. Qwen-T and Qwen-N are Qwen3.8-27B with
  thinking on and off.}
  \label{tab:skills}
  \begin{tabular}{@{}lrrrrrrrrrrr@{}}
\toprule
Skill tasks (cells) & Sol & Opus & Sonnet & Terra & Haiku & Luna & Qwen-T & MiniMax & DeepSeek & Inkling & Qwen-N \\
\midrule
\multicolumn{12}{@{}l}{\textit{Main tasks}} \\
General skills (19) & \textbf{61.0} & 52.4 & 50.5 & 27.7 & 7.6 & 1.3 & 4.9 & 2.1 & 8.4 & 2.7 & 3.5 \\
Whole programs (35) & \textbf{36.3} & 21.4 & 35.4 & 13.2 & 2.8 & 1.0 & 1.4 & 1.4 & 0.9 & 1.3 & 1.7 \\
Function completion (20) & 0.6 & 1.1 & \textbf{1.2} & 1.1 & 0.4 & 0.2 & 0.4 & 1.1 & 1.0 & 0.5 & 0.4 \\
HumanEval functions (3) & 0.0 & 2.1 & 2.1 & \textbf{3.2} & 0.0 & 1.0 & 1.0 & 0.9 & 1.0 & 2.1 & 0.5 \\
\midrule
\multicolumn{12}{@{}l}{\textit{Variants}} \\
Hand design (4) & \textbf{37.2} & 14.6 & 6.6 & 0.4 & 1.1 & 2.0 & 10.0 & 4.8 & 1.3 & 0.6 & 0.2 \\
One-step (4) & \textbf{44.3} & 11.4 & 38.1 & 11.9 & 14.1 & 3.0 & 4.7 & 0.0 & 0.0 & 0.2 & 0.0 \\
Search (4) & \textbf{72.2} & 69.0 & 50.1 & 38.5 & 0.0 & 0.0 & 2.5 & 0.6 & 4.8 & 0.7 & 11.8 \\
Worked examples (3) & 64.7 & 66.4 & \textbf{67.6} & 41.1 & 6.2 & 6.9 & 15.0 & 4.0 & 0.0 & 10.8 & 0.0 \\
Second model (2) & \textbf{60.1} & 45.5 & 51.1 & 9.5 & 1.8 & 0.0 & 0.0 & 0.5 & 3.6 & 0.8 & 0.0 \\
\bottomrule
\end{tabular}

\end{table}

\Cref{tab:skills} splits the skill tasks into groups. The general skills
are the most tractable: GPT-6.1 Sol, Claude Opus 5.5 and Claude Sonnet 5.5 average above 50
on them, and several skills are close to solved. On new facts, ffmpeg commands and OpenSCAD
models, the three strongest agents each score at least 95, and on SQL, two reach 94.9. Other
skills resist every agent: no agent scores above 16 on the number puzzles, 33 on grade-school
math or 43 on calculus. Writing whole programs in 35 languages lies in between, with GPT-6.1 Sol
(36.3) and Claude Sonnet 5.5 (35.4) well ahead of the rest. On completing functions, in the 20
MultiPL-E languages and the three HumanEval tasks, no agent averages more than 3.2. Only four of the 94 skill tasks
leave every agent at 0, and on 42 the best agent reaches at least 50.

The skill variants show how stricter rules and extra help change the scores (\cref{tab:skills});
we compare each variant with the main tasks of the same skills. The three rule sets are posed on
four skills: instruction following, SQL, tool calling and tic-tac-toe. Under the hand-design
rules, the four strongest agents score much lower: GPT-6.1 Sol falls from 73.4 on the main tasks
to 37.2, Claude Opus 5.5 from 48.9 to 14.6, Claude Sonnet 5.5 from 46.0 to 6.6 and GPT-6.1 Terra
from 26.8 to 0.4. One-step edits computed from activations also score lower for all four. Under
the search rules, which allow trial and error over candidate edits, they score about as well as
on the main tasks or better (GPT-6.1 Sol 72.2, Claude Opus 5.5 69.0). The strongest agents thus
succeed mainly by testing candidate edits with forward passes, rather than by designing an edit
from the weights alone. Of the two kinds of help, giving the agent worked examples raises the
scores of three of the four strongest agents, but not GPT-6.1 Sol's. A second model that already has the skill helps none of the four; Claude Opus 5.5 even
scores lower with it (45.5, against 76.6 on the main tasks). With one run per cell and four or
fewer cells per variant, these comparisons are indicative only.

\subsection{Interpreting Scores}\label{sec:analysis}
\begin{figure}[!tb]
  \centering
  \includegraphics[width=\linewidth]{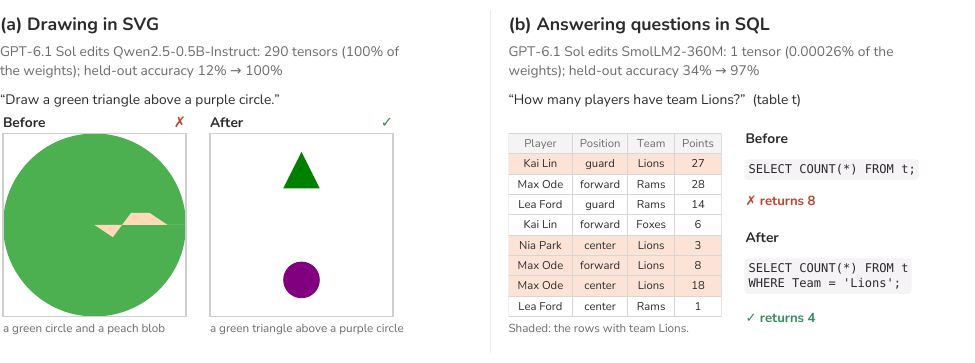}
  \caption{Replies to a practice prompt before and after an edit by GPT-6.1 Sol, for two skills. The header of each
  panel gives the subject model, the size of the edit and the held-out accuracy of the unedited and the edited model.
  \textbf{(a)} The SVG replies, rendered. \textbf{(b)} The SQL queries and the counts they return on table t.}
  \label{fig:sol}
\end{figure}

{\widowpenalty=10000  %
A low score can mean two things: the agent did not edit the model, or its edit did not work. The
harness records how many tensors each submitted model changed, which separates the two. Leaving
out the runs that providers refused, GPT-6.1 Sol edits the model in every edit and skill run, and
Claude Opus 5.5, Claude Sonnet 5.5 and GPT-6.1 Terra hand back the unedited model in at most 7\%
of them. Claude Haiku 5.5 hands it back in 67\% of its runs and GPT-6 Luna in 34\%, and both stop
early, after 15 and 16 replies on average, against 44 to 77 for the four strongest agents. The open
models use as many replies as the strongest agents or more (71 to 91), and MiniMax-M2.7 edits the
model in 98\% of its runs, but 84\% of its edits score 0. Their low scores come from edits that do
not work, not from too few attempts or too small a budget. Thinking matters here too: without it,
Qwen3.8-27B hands back the unedited model in 38\% of its runs instead of 20\%. In short, the weaker closed models
often stop without editing, while the open models edit persistently but rarely in a way that
works: the former most need to attempt an edit, and the latter to make their edits work.
Examples of successful edits by GPT-6.1 Sol are in \cref{fig:sol}; the project page shows before-and-after
examples for every agent on five tasks.\footnote{\url{https://moreirap12.github.io/OpenWSP/}}\par}

\section{Conclusion}\label{sec:conclusion}
We introduced \bench, a benchmark of 152 tasks that measures how well LLMs program the weights of
small open language models: working out how a model was made, editing a specified behaviour, and
teaching a skill, all without gradients. Across the eleven agents we evaluated, the index ranges
from 4.1 to 51.0. The strongest agents already solve tasks that require understanding the weights,
such as finding transplanted layers or removing a backdoor, and on some skills, such as teaching
new facts or answer set programming, they close most of the gap to a perfect score. Two areas leave
clear room for progress: completing functions, where no agent averages more than 3.2, and the open
models, which edit often but rarely with success. As future work, we propose an
experiment in which a model applies WSP to its own weights, for example to improve a skill on
which it is weak; \bench's skill tasks give a way to score the result. We also propose extending \bench to other
architectures and modalities, from world models and vision-language-action models to audio and image flow-matching
models.

\subsection{Limitations}\label{sec:limitations}
The subject models are small (360M--1.7B parameters), and we do not know how the results change
with model size. Each agent runs each cell once, so the numbers for single classes and variants
rest on few runs. When a provider's safety filter refused to answer, the run scored 0; this affects 28 of Claude
Haiku 5.5's runs, so its index partly reflects the filter (\cref{sec:setup}). The agents run
through different interfaces (Claude Code, Codex and a plain chat endpoint for the open models),
so each result describes a model together with its interface. Finally, the held-out problems of
the HumanEval coding skills and grade-school math come from public datasets, so an agent could in
principle write their answers into the model.

\bibliographystyle{plainnat}
\bibliography{references}

\clearpage
\appendix
\crefalias{section}{appendix}  %

\section{Task Details}\label{app:tasks}

\Cref{tab:general-skills} lists the 18 general skills and \cref{tab:coding-skills} the 58 coding
skills. For every skill, the agent edits the subject model, and the checker scores the edited
model's replies to the held-out prompts.

\begin{table}[!htbp]
  \centering
  \small
  \caption{The 18 general skills.}
  \label{tab:general-skills}
  \begin{tabularx}{\linewidth}{@{}>{\raggedright\arraybackslash}p{2.6cm}>{\raggedright\arraybackslash}p{2.7cm}>{\raggedright\arraybackslash}X>{\raggedright\arraybackslash}X@{}}
    \toprule
    Skill & Subject model & The edited model must & Checked by \\
    \midrule
    Instruction following & Qwen2.5-0.5B; Qwen3-0.6B-Base & follow verifiable formatting instructions (for example all lowercase, or exactly three bullet points) on a given topic & rule checks and a topic check \\
    Tool calling & Qwen2.5-0.5B & answer a request with a JSON list of function calls & match with the reference calls \\
    SQL & SmolLM2-360M & translate a question about a table into a query & run in SQLite; same rows as the reference \\
    Spreadsheet formulas & Qwen2.5-0.5B-Instruct & write one Excel formula over given rows & evaluated by a formula engine \\
    Calculus & Qwen3-0.6B-Base & differentiate or integrate a function & symbolic comparison \\
    Grade-school math & Qwen3-0.6B-Base & solve a word problem & final number \\
    Number puzzles & Qwen2.5-0.5B-Instruct & combine given numbers with $+$, $-$, $\times$, $\div$ to reach a target & each number used once; equals the target \\
    New facts & Qwen2.5-0.5B-Instruct & recall a set of fictional facts, asked directly and in paraphrase & the right value in the reply \\
    Constrained writing & Qwen2.5-0.5B-Instruct & write acrostics, stories with an exact number of sentences that avoid a letter, and rhyming poems & every constraint holds \\
    Answer set programming & Qwen3-1.7B & write a clingo program for a small combinatorial problem & same answer sets as the reference \\
    SVG drawing & Qwen2.5-0.5B-Instruct & draw two coloured shapes in a given spatial relation & shapes, colours and relation \\
    3D modelling & Qwen3-1.7B & write an OpenSCAD program for a described part & volume, bounding box and position of the part \\
    Video editing & Qwen2.5-0.5B-Instruct & write one ffmpeg command, for example to trim, resize or change the frame rate & properties of the output video \\
    LEGO building & Qwen3-1.7B & list the bricks that build a specified structure & overlap with the target structure \\
    Slide writing & Qwen2.5-0.5B-Instruct & write a Marp slide deck under format rules & share of rules met \\
    Slide design & Qwen2.5-0.5B-Instruct & design a Marp slide deck under design rules & rules checked on the rendered deck \\
    Tic-tac-toe & Qwen3-1.7B-Base & play full games against a tactical opponent & game results \\
    Connect-4 & Qwen3-1.7B-Base & play full games against a tactical opponent & game results \\
    \bottomrule
  \end{tabularx}
\end{table}

\begin{table}[!htbp]
  \centering
  \small
  \caption{The 58 coding skills, in three groups, covering 48 programming languages.}
  \label{tab:coding-skills}
  \begin{tabularx}{\linewidth}{@{}>{\raggedright\arraybackslash}p{2.6cm}>{\raggedright\arraybackslash}p{2.7cm}>{\raggedright\arraybackslash}X@{}}
    \toprule
    Group & Subject model & Task, checker and languages \\
    \midrule
    Whole programs (35) & Qwen2.5-1.5B-Instruct & write a complete program for a small task; it is compiled if needed and run on four test inputs. Ada, ALGOL~68, AWK, BASIC, C, COBOL, Common Lisp, Crystal, Erlang, Forth, Fortran, Go, GolfScript, Haskell, Haxe, Java, Kotlin, Lua, Nim, OCaml, Octave, Pascal, Perl, Pike, Prolog, Python, R, Ruby, Rust, Scheme, Shakespeare, SQL, Standard ML, Tcl, Vala \\
    \addlinespace
    Function completion (20) & Qwen2.5-0.5B & continue a function signature and doc comment from MultiPL-E's translation of HumanEval; the problem's unit tests must pass. Bash, C++, C\#, Clojure, D, Elixir, Go, Haskell, Java, Julia, OCaml, Perl, PHP, R, Racket, Ruby, Rust, Scala, Swift, TypeScript \\
    \addlinespace
    HumanEval functions (3) & Qwen2.5-0.5B-Instruct; Qwen2.5-0.5B for Python & complete a HumanEval function, given in a chat message (as plain text for Python); the unit tests must pass. JavaScript, Lua, Python \\
    \bottomrule
  \end{tabularx}
\end{table}

\section{Tool Limits}\label{app:prompts}
One code execution may take up to 270 seconds, and its printed output is cut at 10,000
characters. A forward-pass request can ask for up to 512 generations at once, and a request that
fails does not use the budget.

\section{Additional Results}\label{app:results}

\Cref{tab:main} gives the main results in full, and \cref{tab:per-class} gives every agent's mean score in each task class (\cref{tab:classes}).

\begin{table}[!htbp]
  \centering
  \small
  \caption{Each agent's index is its mean score over the 168 cells (0 = baseline, 100 = perfect), with a 95\% bootstrap confidence interval; Read, Edit and Skill are the same mean
  over each family's cells. \emph{Solved} counts cells with a perfect score (100 after rounding), and \$/task is the
  mean API list price of one run. Each agent ran each cell once; refused runs score 0.}
  \label{tab:main}
  \begin{tabular}{@{}llr rrr rr@{}}
\toprule
Agent & Setting & Index {\scriptsize [95\% CI]} & Read & Edit & Skill & Solved & \$/task \\
 & & {\scriptsize 168 cells} & {\scriptsize 43} & {\scriptsize 31} & {\scriptsize 94} & & \\
\midrule
\multicolumn{8}{@{}l}{\textit{Closed}} \\
GPT-6.1 Sol & high & \textbf{51.0} {\scriptsize [45.2, 57.1]} & \textbf{74.1} & \textbf{65.2} & \textbf{35.8} & \textbf{26} & 0.61 \\
Claude Opus 5.5 & high & 37.6 {\scriptsize [32.1, 43.3]} & 53.0 & 51.4 & 26.0 & 13 & 4.41 \\
Claude Sonnet 5.5 & high & 33.7 {\scriptsize [28.3, 39.3]} & 28.8 & 49.0 & 31.0 & 8 & 1.89 \\
GPT-6.1 Terra & high & 29.0 {\scriptsize [23.5, 34.6]} & 53.3 & 38.9 & 14.5 & 14 & 0.12 \\
Claude Haiku 5.5 & high & 8.9 {\scriptsize [5.5, 12.4]} & 13.0 & 19.8 & 3.5 & 1 & 0.03 \\
GPT-6 Luna & high & 6.8 {\scriptsize [3.9, 10.1]} & 16.5 & 10.3 & 1.1 & 1 & 0.01 \\
\midrule
\multicolumn{8}{@{}l}{\textit{Open weights}} \\
Qwen3.8-27B & thinking & 7.5 {\scriptsize [4.6, 10.8]} & 17.1 & 8.3 & 2.9 & 1 & 0.83 \\
MiniMax-M2.7 & thinking & 4.5 {\scriptsize [2.1, 7.1]} & 13.8 & 0.2 & 1.6 & 2 & 0.18 \\
DeepSeek-V4-Flash & thinking, high & 4.4 {\scriptsize [2.5, 6.5]} & 8.5 & 3.9 & 2.6 & 0 & 0.06 \\
Inkling 2 Mini & thinking, high & 4.1 {\scriptsize [2.2, 6.4]} & 7.8 & 6.6 & 1.6 & 0 & -- \\
Qwen3.8-27B & no thinking & 4.1 {\scriptsize [2.2, 6.6]} & 9.6 & 3.0 & 2.0 & 0 & 0.18 \\
\bottomrule
\end{tabular}

\end{table}

\begin{table}[!htbp]
  \centering
  \footnotesize
  \setlength{\tabcolsep}{3.5pt}
  \caption{Mean score per task class, with the number of cells in each class. Agents are
  ordered as in \cref{tab:main}; Qwen-T and Qwen-N are Qwen3.8-27B with
  thinking on and off.}
  \label{tab:per-class}
  \begin{tabular}{@{}lrrrrrrrrrrr@{}}
\toprule
Class (cells) & Sol & Opus & Sonnet & Terra & Haiku & Luna & Qwen-T & MiniMax & DeepSeek & Inkling & Qwen-N \\
\midrule
\multicolumn{12}{@{}l}{\textit{Read}} \\
O2 (6) & \textbf{63.2} & 56.0 & 53.9 & 52.0 & 0.0 & 0.0 & 0.0 & 0.0 & 0.0 & 0.0 & 0.0 \\
O3 (3) & 31.8 & 10.4 & 3.2 & 16.5 & 5.7 & \textbf{41.1} & 23.1 & 33.2 & 9.9 & 5.8 & 19.3 \\
P1 (3) & \textbf{75.8} & 35.9 & 56.3 & 61.6 & 36.8 & 62.2 & 32.6 & 39.3 & 34.6 & 44.2 & 56.6 \\
P2 (3) & \textbf{100.0} & \textbf{100.0} & 0.0 & \textbf{100.0} & 0.0 & 5.6 & 0.0 & 0.0 & 0.0 & 0.0 & 16.0 \\
P3 (6) & \textbf{83.8} & 64.6 & 61.2 & 83.2 & 26.8 & 37.5 & 28.0 & 21.6 & 11.9 & 21.1 & 9.1 \\
P3o (4) & \textbf{62.3} & 25.0 & 25.0 & 50.0 & 25.0 & 0.0 & 25.0 & 25.0 & 0.0 & 0.0 & 0.0 \\
S2 (4) & 49.2 & \textbf{66.5} & 5.0 & 42.2 & 10.0 & 0.0 & 31.0 & 16.0 & 0.0 & 0.0 & 21.0 \\
S3 (4) & \textbf{97.0} & 64.0 & 31.5 & 49.9 & 0.0 & 40.0 & 44.0 & 21.0 & 40.0 & 15.0 & 0.0 \\
S6 (10) & \textbf{84.8} & 49.6 & 12.2 & 37.9 & 13.1 & 0.0 & 0.0 & 0.0 & 0.0 & 0.0 & 0.0 \\
\midrule
\multicolumn{12}{@{}l}{\textit{Edit}} \\
W1 (1) & 97.9 & \textbf{100.0} & \textbf{100.0} & 93.2 & 0.0 & 0.0 & 0.0 & 0.0 & 6.4 & 0.0 & 0.0 \\
W2 (1) & 44.9 & 42.3 & 0.0 & \textbf{71.8} & 0.0 & 5.1 & 0.0 & 0.0 & 9.0 & 0.0 & 0.0 \\
W3 (1) & 15.4 & 0.0 & 17.9 & \textbf{25.6} & 9.0 & 9.0 & 9.0 & 0.0 & 7.7 & 9.0 & 0.0 \\
W4 (10) & \textbf{98.4} & 69.2 & 52.8 & 64.4 & 43.3 & 7.7 & 15.5 & 0.0 & 9.5 & 13.7 & 8.3 \\
W4x (12) & 44.7 & \textbf{60.1} & 49.3 & 26.2 & 12.2 & 17.9 & 6.2 & 0.4 & 0.3 & 1.1 & 0.5 \\
W5 (6) & \textbf{56.8} & 6.4 & 46.9 & 9.7 & 3.9 & 2.3 & 3.3 & 0.0 & 0.0 & 7.6 & 0.5 \\
\midrule
\multicolumn{12}{@{}l}{\textit{Skill}} \\
K (94) & \textbf{35.8} & 26.0 & 31.0 & 14.5 & 3.5 & 1.1 & 2.9 & 1.6 & 2.6 & 1.6 & 2.0 \\
\bottomrule
\end{tabular}

\end{table}

\end{document}